\documentclass{article}

\usepackage{docmute}

\usepackage{arxiv}

\usepackage[utf8]{inputenc}
\usepackage[T1]{fontenc}

\usepackage{amsmath}
\usepackage{amssymb}
\usepackage{amsfonts}
\usepackage{amsthm}
\usepackage{bm}
\usepackage{comment}
\usepackage{fancybox}
\usepackage{framed}
\usepackage{color}
\usepackage[dvipdfmx]{graphicx}
\usepackage{multicol}
\usepackage{multirow}
\usepackage{pdflscape}
\usepackage{verbatim}
\usepackage{subfigure}
\usepackage{cite}
\usepackage{hyperref}
\usepackage{url}

\hypersetup{
    colorlinks = true,
    urlcolor = blue,
    linkcolor = red,
    citecolor = green
}

\usepackage{algorithm}
\usepackage[noend]{algpseudocode}
\usepackage{algorithmicx}

\expandafter\def\expandafter\UrlBreaks\expandafter{\UrlBreaks
  \do\a\do\b\do\c\do\d\do\e\do\f\do\g\do\h\do\i\do\j%
  \do\k\do\l\do\m\do\n\do\o\do\p\do\q\do\r\do\s\do\t%
  \do\u\do\v\do\w\do\x\do\y\do\z\do\A\do\B\do\C\do\D%
  \do\E\do\F\do\G\do\H\do\I\do\J\do\K\do\L\do\M\do\N%
  \do\O\do\P\do\Q\do\R\do\S\do\T\do\U\do\V\do\W\do\X%
  \do\Y\do\Z}

\algrenewcommand\algorithmicindent{1.0em}

\algnewcommand\algorithmicforeach{\textbf{for each}}
\algdef{S}[FOR]{ForEach}[1]{\algorithmicforeach\ #1\ \algorithmicdo}

\algnewcommand\AlgAnd{\textbf{and} }
\algnewcommand\AlgOr{\textbf{or} }

\algrenewcommand\textproc{}

\algnewcommand{\Initialize}[1]{
	\State \textbf{Initialize:}
 	\State \hspace*{\algorithmicindent}\parbox[t]{0.8\linewidth}{\raggedright #1}}

\title{Federated Learning of Neural ODE Models with Different Iteration Counts}

\author{
  Yuto Hoshino\\
  Keio University\\
  3-14-1 Hiyoshi, Kohoku-ku, Yokohama, Japan\\
  \texttt{hoshino@arc.ics.keio.ac.jp}\\
  \And
  Hiroki Kawakami\\
  Keio University\\
  3-14-1 Hiyoshi, Kohoku-ku, Yokohama, Japan\\
  \texttt{kawakami@arc.ics.keio.ac.jp}\\
  \And
  Hiroki Matsutani \\
  Keio University\\
  3-14-1 Hiyoshi, Kohoku-ku, Yokohama, Japan\\
  \texttt{matutani@arc.ics.keio.ac.jp} \\
}

\begin{document}

\maketitle

\begin{abstract}
Federated learning is a distributed machine learning approach in which
clients train models locally with their own data and upload them to a
server so that their trained results are shared between them without
uploading raw data to the server.
There are some challenges in federated learning, such as communication
size reduction and client heterogeneity.
The former can mitigate the communication overheads, and the latter
can allow the clients to choose proper models depending on their
available compute resources.
To address these challenges, in this paper, we utilize Neural ODE
based models for federated learning.
The proposed flexible federated learning approach can reduce the
communication size while aggregating models with different iteration
counts or depths.
Our contribution is that we experimentally demonstrate that the
proposed federated learning can aggregate models with different
iteration counts or depths.
It is compared with a different federated learning approach in terms
of the accuracy.
Furthermore, we show that our approach can reduce communication size
by up to 92.4\% compared with a baseline ResNet model using CIFAR-10
dataset.
\end{abstract}




\section{Introduction} \label{sec:intro}
In traditional cloud-based machine learning systems, sending personal
data to cloud servers has become problematic from a privacy
perspective.
Federated learning \cite{Federated} is a distributed machine learning
approach that can keep privacy-sensitive raw data decentralized.
In the federated learning, clients receive a model from the server.
Then they train the model with their own data and upload trained
parameters to the server.
The server aggregates the trained parameters received from the clients
and sends back the aggregated parameters to the clients.
These steps are repeated until the training process is converged.
This eliminates the need to upload privacy-sensitive raw data to the
server.

However, there are some challenges in the federated learning, such as
communication size reduction and client heterogeneity.
Communication size affects communication delay and power
consumption of clients.
It is affected by the machine learning model size.
Regarding the client heterogeneity, not all clients always have the
same hardware, compute resources, or training data.
A client may use a deeper model for high accuracy, while another
client may use a shallower model to reduce the computation cost.
In this paper, we exploit Neural ODE \cite{odenet} as a federated
learning model to address these challenges.

For image recognition tasks, one of methods to improve accuracy is
increasing the number of convolutional layers to build a deeper neural
network.
ResNet \cite{resnet} is one of well-known CNN models that stack many
residual blocks that contain convolutional layers and shortcut
connections.
Neural ODE utilizes a similarity to ODE (Ordinary Differential
Equation) to implement deep neural networks consisting of residual
blocks.
Since it can be approximated to a ResNet model by repeatedly using the
same weight parameters, it can reduce the weight parameters.
In addition, it can be approximated to ResNet models with different
depths by changing the iteration counts without increasing the number
of parameters.
dsODENet \cite{dsode} is a lightweight model that combines the ideas
of Neural ODE and depthwise separable convolution \cite{mobilenets} to
further reduce the parameter size and computation cost.
These Neural ODE models are smaller than ResNet.

In this paper, we introduce a flexible federated learning that allows
clients to use models with different iteration counts and reduces the
communication size by using Neural ODE based models.
Our contributions are listed below
\footnote{An early stage of this work appeared in our workshop paper
\cite{hoshinocw}.
In this paper, we experimentally demonstrate that the proposed
approach can aggregate models with different iteration counts.
It is compared to a federated learning approach that uses knowledge
distillation.}.
\begin{itemize}
\item We propose to use the Neural ODE models for federated learning so
  that a server can aggregate models with different iteration counts.
  This can enhance the client heterogeneity since clients can use
  models with different iteration counts.
  In addition, using the Neural ODE models can significantly reduce
  the communication size.
\item We experimentally demonstrate that the proposed flexible
  federated learning can aggregate these models with different
  iteration counts.
  It is compared with a federated learning approach that uses
  knowledge distillation.
  We discuss the pros and cons of the proposed approach.
\end{itemize}

The rest of this paper is organized as follows.
Section \ref{sec:related} introduces baseline technologies behind our
proposal.
Section \ref{sec:design} proposes the flexible federated learning
approach and shows the feasibility of the proposed approach.
Section \ref{sec:eval} evaluates the proposed approach in terms of the
accuracy and communication size.
Section \ref{sec:conc} concludes this paper.


\section{Related Work} \label{sec:related}
\subsection{Federated Learning}
Federated Averaging (FedAvg) is a basic federated learning algorithm
proposed in \cite{Federated}.
Algorithm \ref{alg:federated-averaging} shows the server- and
client-side flows, where $K$ is the total number of clients, $k$
is their index, and $\mathcal{P}_k$ is data at client $k$.
Also, $B$ is the size of a local mini-batch, $E$ is the number of
epochs to be trained by each client, and $\eta$ is a given learning
rate.
In this algorithm, the first step is to initialize global weight
parameters of the model.
Then, $m$ clients are randomly selected from $K$ clients, and the
server sends the global parameters to the selected clients.
The size of $m$ is determined by a client fraction parameter $C$.
The weight parameters are updated at each epoch ($E$ epochs in total)
by each client based on the formula in line \ref{alg:update}.
After $E$ updates, the clients send their trained local parameters to
the server.
The server aggregates the received local parameters by taking the
average based on the formula in line \ref{alg:average}, where $n$ is
the total number of data and $n_k$ is the total number of data at
client $k$.
The aggregated parameters are then sent back to the clients as global
parameters.
The above steps are repeated $t$ rounds.

\begin{algorithm}[t]
    \caption{Federated Averaging \cite{Federated}}
    \label{alg:federated-averaging}
    \begin{algorithmic}[1]
        \Function{ExecuteServer}{\null}
            \State Initialize $w_0$
            \ForEach{round $t = 1, 2, \ldots$}
                \State $m \gets \max(r \cdot K, 1)$
                \State $S_t \gets \text{(random set of $m$ clients)}$
                \ForEach{client $k  \in S_t$ in parallel}
                    \State $w_{t + 1}^k \gets$ \Call{ClientUpdate}{$k, w_t$}
                \EndFor
                \State $w_{t + 1} \gets \displaystyle \sum_{k = 1}^K \frac{n_k}{n} w_{t + 1}^k$ \label{alg:average}
            \EndFor
        \EndFunction

        \Function{ClientUpdate}{$k, w$} \hfill $\triangleright$ \textit{Run on client $k$}
            \State $\mathcal{B} \gets \text{(split $\mathcal{P}_k$ into batches of size $B$)}$
            \ForEach{local epoch $i$ from $1$ to $E$}
                \ForEach{batch $b \in \mathcal{B}$}
                    \State $w \gets w - \eta \nabla \ell(w; b)$ \label{alg:update}
                \EndFor
            \EndFor
        \EndFunction
    \end{algorithmic}
\end{algorithm}

Many federated learning technologies have been studied since FedAvg
was proposed in 2016.
These technologies are surveyed in \cite{survey}.
Data heterogeneity is one of important research challenges in these
studies since it is a major cause of accuracy degradation.
For instance, since a local model is optimized toward the local optima
by the client, it may be distant from other clients.
Thus, their averaged global model may be far from a part of clients.
To deal with this problem, FedProx \cite{fedprox} uses an additional
proximal term to limit the number of local updates, and SCAFFOLD
\cite{scaf} uses a variance reduction to correct local updates.
These algorithms aim to improve the local training step of FedAvg.
In contrast, Personalized Federated Averaging \cite{perfed} and
Adaptive Personalized Federated Learning \cite{apfl} aim to make
personalized models that can achieve good accuracy in local clients.

Another challenge is the clients' model heterogeneity.
Since not all clients always have the same compute resources,
selecting a proper model for each client can help the client
heterogeneity.
FedFeNN \cite{fedhenn} and FedDF \cite{feddf} address the model
heterogeneity.
In FedHeNN, each client trains its own model but pulls the
representations learned by different clients closer by adding a
proximal term to the client's loss function \cite{fedhenn}.
FedDF is a federated learning algorithm that utilizes a knowledge
distillation at the model aggregation step of the federated learning
server.
In the knowledge distillation step at the server, instead of averaging
local parameters received from clients, a batch of sample data is
predicted by the local parameters and their output logits are
averaged.
The averaged logits are then used for updating the client models at
the server.
Although FedDF allows a federated leaning of different model
architectures, model training is needed at the aggregation step of the
server in addition to local training at the clients.
This means that all the client models joining the federated learning
are also needed at the server.

\subsection{ResNet and Neural ODE}
ResNet \cite{resnet} is a well-known neural network architecture that
can increase the number of stacked layers or building blocks by
introducing shortcut connections.
Using a shortcut connection, an input feature map to a building block
is temporarily saved, and then it is added to the original output of
the building block to generate the final output of the block.
In this paper, one building block in ResNet is called ResBlock.

ODE is composed of an unknown function and its ordinary derivatives.
To obtain an approximate numerical solution, an ODE solver such as the
first-order Euler method and higher-order Runge-Kutta method can be
used.
Based on a similarity between the network structure with shortcut
connections and the ODE solver, one building block can be interpreted
as one step in the ODE solver as suggested in \cite{odenet}.
Assuming that the Euler method is used as an ODE solver, it can be
interpreted that the first-order approximation is applied to solve an
output of the building block.
In this paper, one building block is called ODEBlock, and the whole
network architecture consisting of ODEBlocks is called ODENet.

\subsection{Depthwise Separable Convolution}\label{ssec:dsc}
CNN is composed of multiple layers, such as convolutional layers,
pooling layers, and fully-connected layers.
Although CNN achieves a good accuracy in image recognition tasks, each
convolutional layer has many parameters.
Let $N$, $M$, and $N_K$ be the number of input channels, the number of
output channels, and the kernel size of one side, respectively.
The number of parameters in one convolutional layer is $NMN_K^2$.

Depthwise separable convolution \cite{mobilenets} divides this
convolutional layer into two convolutional steps: depthwise
convolutional step and pointwise convolutional step.
In the depthwise convolutional step, a convolutional operation
involving only spatial direction (the size is $N_K^2$) is applied for
each input feature map.
Different weight parameters are used for each of $N$ input channels;
thus its weight parameter size is $NN_K^2$.
Then, an output feature map of the depthwise convolutional step is fed
to the pointwise convolutional step as an input.
A $1 \times 1$ convolutional operation is applied for each input
feature map and for each output channel; thus its weight parameter
size is $NM$.
The weight parameter size of the depthwise separable convolution is
$NN_K^2+NM$ in total, which is approximately $N_K^2$ times reduction,
assuming that $N, M \gg N_K$.

As a low-cost CNN model, dsODENet \cite{dsode} applies the depthwise
separable convolution to convolutional layers of ODEBlocks in order to
further reduce the parameter size.
It was originally proposed to be implemented on resource-limited FPGA
(Field-Programmable Gate Array) devices \cite{dsode}.
In this paper, we use dsODENet as a federated learning model
architecture in addition to ODENet.
Their detailed structures are illustrated in the next section.


\section{Proposed Federated Learning} \label{sec:design}


\begin{figure}[htb]
    \begin{minipage}{0.49\textwidth}
            \begin{center}
            \includegraphics[height=60mm]{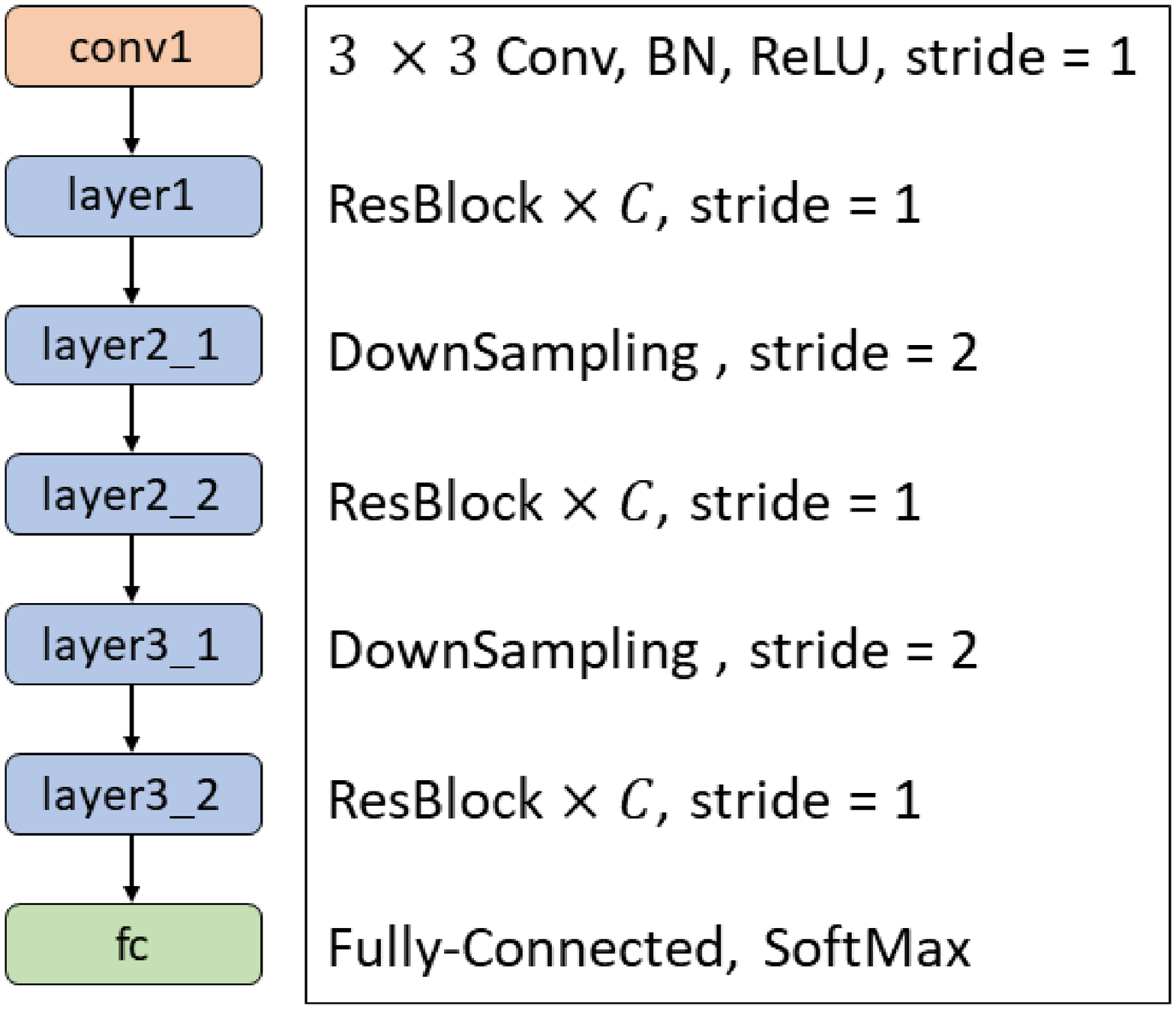}
            \caption{Structure of 7-block ResNet}
            \label{fig:resnet-struc}
            \end{center}
    \end{minipage}
    \begin{minipage}{0.49\textwidth}
            \begin{center}
            \includegraphics[height=60mm]{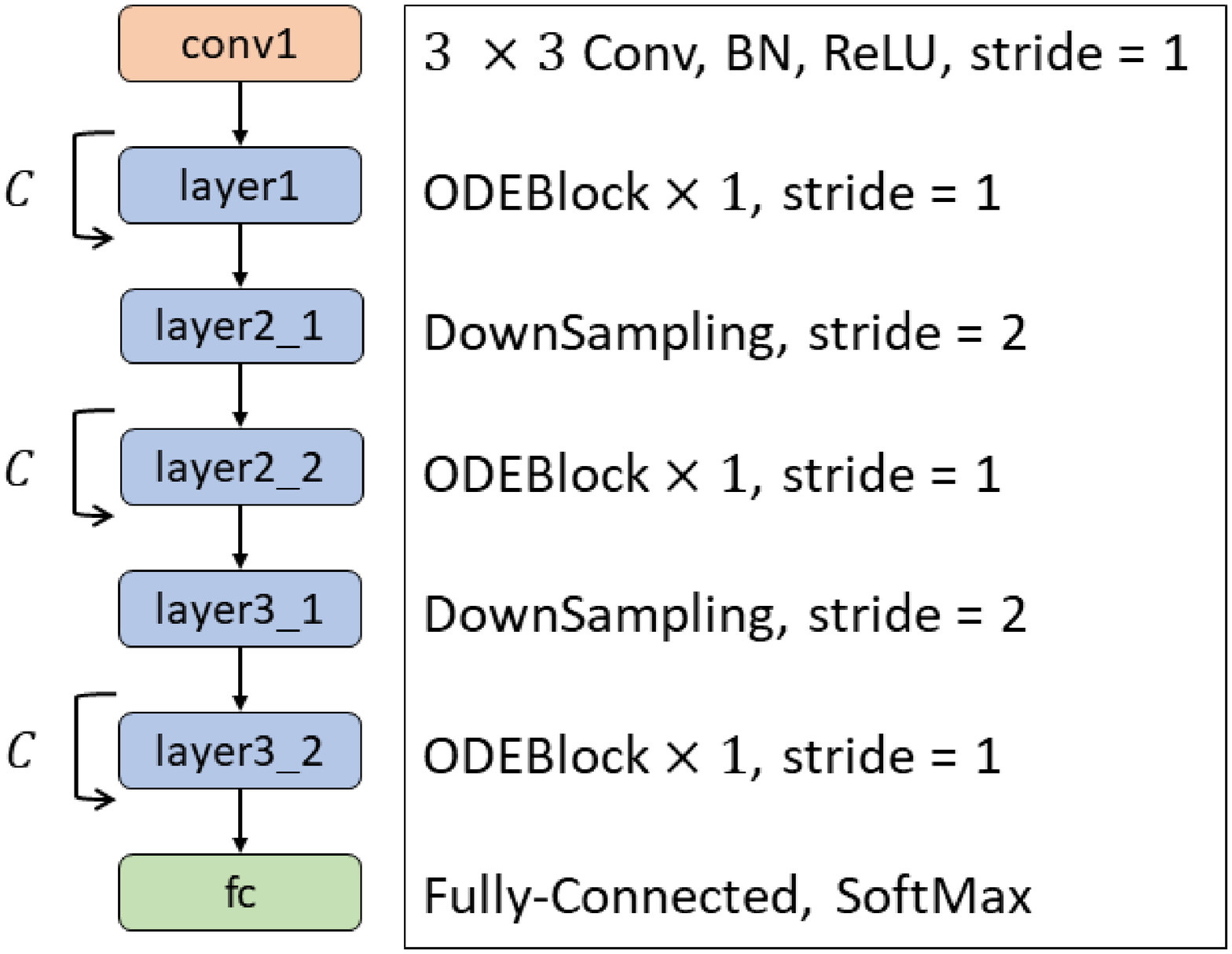}
            \caption{Structure of 7-block ODENet}
            \label{fig:odenet-struc}
            \end{center}
    \end{minipage}
\end{figure}

In the traditional federated learning such as FedAvg, the server and
all clients have to use the same model.
For example, models with different layer depths cannot be averaged
when ResNet is used as a model architecture.
However, in a real environment, client devices are not the same and
are likely to have different compute resources, such as memory
capacity and computation power.
Since the traditional federated learning cannot aggregate models with
different depths, a common model used by all the clients should be
carefully selected.
In addition, it is necessary to reduce the communication size involved
in exchanging weight parameters between the server and clients.
In this paper, we use ODENet and dsODENet to enable a flexible
federated learning between models with different layer iteration
counts and significantly reduce the communication size.
These target models are illustrated in Section \ref{ssec:models}.
We discuss the feasibility of the proposed federated learning in
Sections \ref{ssec:comp} and \ref{ssec:diffed}.

\subsection{Target Models}\label{ssec:models}
In this section, we first illustrate the structures and sizes of
ResNet and ODENet.
Then, we illustrate dsODENet.

Figures \ref{fig:resnet-struc} and \ref{fig:odenet-struc} show basic
structures of ResNet and the corresponding ODENet.
They consist of seven blocks including conv1 and fc.
In the ResNet model, conv1 performs convolutional operations as a
pre-processing layer, and fc is a post-processing fully-connected layer.
After the conv1, $C$ physically-stacked ResBlocks are executed
in block1.
block2\_1 is a downsampling ResBlock to reduce the feature map size,
and $C$ physically-stacked ResBlocks on the output of block2\_1 are
executed in block2\_2.
The same operation is performed for block3\_1 and block3\_2 too.


ODENet replaces ResBlocks in Figure \ref{fig:resnet-struc} with
ODEBlocks as shown in Figure \ref{fig:odenet-struc}.
In ResNet, $C$ ResBlocks are physically-stacked in block1, block2\_2,
and block3\_2, while ODENet replaces these $C$ ResBlocks with a single
ODEBlock.
Instead, ODEBlock is executed $C$ times in block1, block2\_2, and
block3\_2.
A downsampling ODEBlock is executed only once in block2\_1 and
block3\_1, respectively.

Here, we analyze the numbers of parameters of ResNet and ODENet.
Let $O(L)$ be the number of parameters in one ResBlock and ODEBlock.
In ResNet, $C$ ResBlocks are stacked in one block, so the number of
parameters is $O(CL)$.
In contrast, ODENet repeats ODEBlock $C$ times in one block, so the
number of parameters is $O(L)$.
The parameter size reduction by ODENet becomes large as $C$ is
increased.
As shown, the communication size can be reduced by using ODENet as a
federated learning model instead of ResNet.


\begin{figure}
    \centering
    \includegraphics[width=0.4\linewidth]{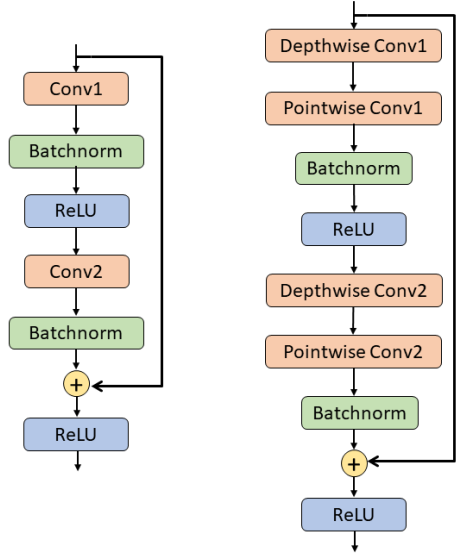}
    \caption{Structures of ODEBlock and dsODEBlock in ODENet and dsODENet}
    \label{fig:odeblock}
\end{figure}

\begin{table}[t]
\begin{minipage}{0.49\textwidth}
	\caption{Accuracy of ODENet models trained and tested with
          same or different depths}
	\centering
	\begin{tabular}{|r|r|r|}
	  \hline
		Trained as & Teseted as & Accuracy [\%] \\
	  \hline
	    ODENet-34 & ODENet-34 & 75.46 $\pm$ 0.41\\
				 & ODENet-50 & 75.36 $\pm$ 0.43\\
				 & ODENet-101 & 75.26 $\pm$ 0.46\\
				 \hline 
		ODENet-50  & ODENet-34 & 76.13 $\pm$ 0.42\\
				   & ODENet-50 & 76.02 $\pm$ 0.42\\
					& ODENet-101 & 75.91 $\pm$ 0.52\\
		\hline  
		ODENet-101  & ODENet-34 & 76.25 $\pm$ 0.45\\
				   & ODENet-50 & 76.14 $\pm$ 0.47\\
					& ODENet-101 & 76.13 $\pm$  0.45\\
		\hline  
	\end{tabular}
	\label{table:odeDifferent}
\end{minipage}
\begin{minipage}{0.49\textwidth}
	\caption{Accuracy of dsODENet models trained and tested with
          same or different depths}
	\centering
	\begin{tabular}{|r|r|r|}
	  \hline
		Trained as & Tested as & Accuracy [\%]\\
	  \hline
	  dsODENet-34 & dsODENet-34 & 71.67 $\pm$ 0.53\\
				 & dsODENet-50 & 71.74 $\pm$ 0.68\\
				 & dsODENet-101 & 71.45 $\pm$ 0.68\\
				 \hline  
		dsODENet-50  & dsODENet-34 &72.24 $\pm$ 0.61 \\
				   & dsODENet-50 & 72.10 $\pm$ 0.62\\
					& dsODENet-101 & 72.00 $\pm$ 0.69\\
		\hline 
		dsODENet-101  & dsODENet-34 & 72.28 $\pm$ 0.64\\
				   & dsODENet-50 & 72.11 $\pm$ 0.62\\
					& dsODENet-101 & 72.13 $\pm$ 0.64\\
		\hline  
	\end{tabular}
	\label{table:dsodeDifferent}
\end{minipage}
\end{table}

In addition, the number of parameters can be further reduced by using
dsODENet \cite{dsode} as mentioned in Section \ref{ssec:dsc}.
Figure \ref{fig:odeblock} left shows a structure of an ODEBlock in
ODENet.
Figure \ref{fig:odeblock} right shows that of dsODEBlock in dsODENet,
in which each convolutional layer of the ODEBlock is replaced with two
convolutional steps: the depthwise convolutional step and the
pointwise convolutional step.
Conv1 and Conv2 are convolutional layers, and ReLU (Rectified Linear
Unit) is an activation function.
This modification can reduce the model and communication sizes
compared with ODENet.

\subsection{Weight Compatibility with Different Depths}\label{ssec:comp}
In the case of ResNet, models with different depths have different
numbers of stacked ResBlocks (i.e., different $C$) in their block1,
block2\_2, and block3\_2.
If we use FedAvg for federated learning, basically these different
ResNet models cannot be averaged at the server due to the model
incompatibility.
In the case of ODENet, one ODEBlock is repeated $C$ times in block1,
block2\_2, and block3\_2 of ODENet.
In other words, ODENet models with different depths differ only in the
numbers of iterations of ODEBlocks, not in the number of ODEBlocks.
Therefore, the structure of the ODENet models with different $C$ is
the same, so their parameters can be averaged at the server.
Using ODENet as a federated learning model enables a flexible
federated learning between models with different iteration counts and
fully utilizes the performance of each client device.
dsODENet \cite{dsode} also enables such a flexible federated learning
between models with different iteration counts for the same reason as
ODENet.

Here, we examine if the above observation can work.
This section focuses on the weight parameter compatibility of models
which have different $C$ to demonstrate the feasibility of the
proposed federated learning.
Specifically, inference accuracies of ODENet and dsODENet models which
were trained for the same or different $C$ are evaluated.
Please note that, in the following, we use the total number of
executed convolutional layers (denoted as $N$) to represent the depths
of the ResNet, ODENet, and dsODENet models.
We assume $N=6C+6$ in this experiment \footnote{In Figure
\ref{fig:resnet-struc}, ResBlocks are executed $3C+2$ times, each
contains two convolutional layers.
The pre- and post-processing (conv1 and fc) layers are also included
in $N$; thus $N=2(3C+2)+2$.} and use $N$ = 34, 50, and 101.
For example, the inference accuracy of ODENet-34 is evaluated using
weight parameters trained as ODENet-50.
We selected two models from $N$ = 34, 50, and 101: one for training
and another for inference.
The experiment is performed 100 times for each combination.
CIFAR-10 dataset \cite{cifar10} is used for the training and inference.
The same experiment is also performed for dsODENet.

\begin{figure}[t]
\begin{minipage}{0.49\textwidth}
	\centering
	\includegraphics[width=1\linewidth]{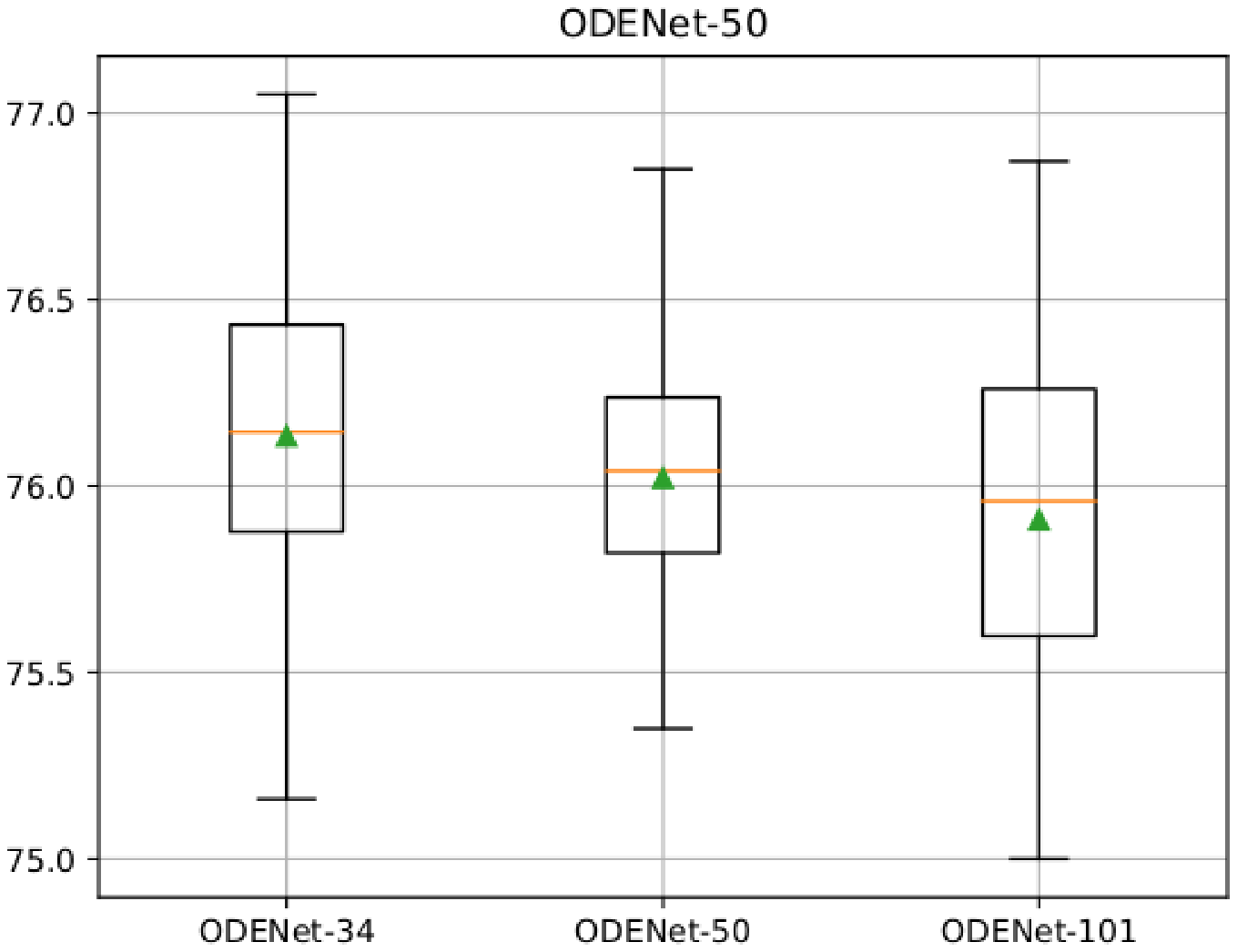}
	\caption{Accuracy of ODENet-34, 50, and 101 using parameters
          trained as ODENet-50 [\%]}
	\label{fig:odeweight50}
\end{minipage}
\begin{minipage}{0.49\textwidth}
	\centering
	\includegraphics[width=1\linewidth]{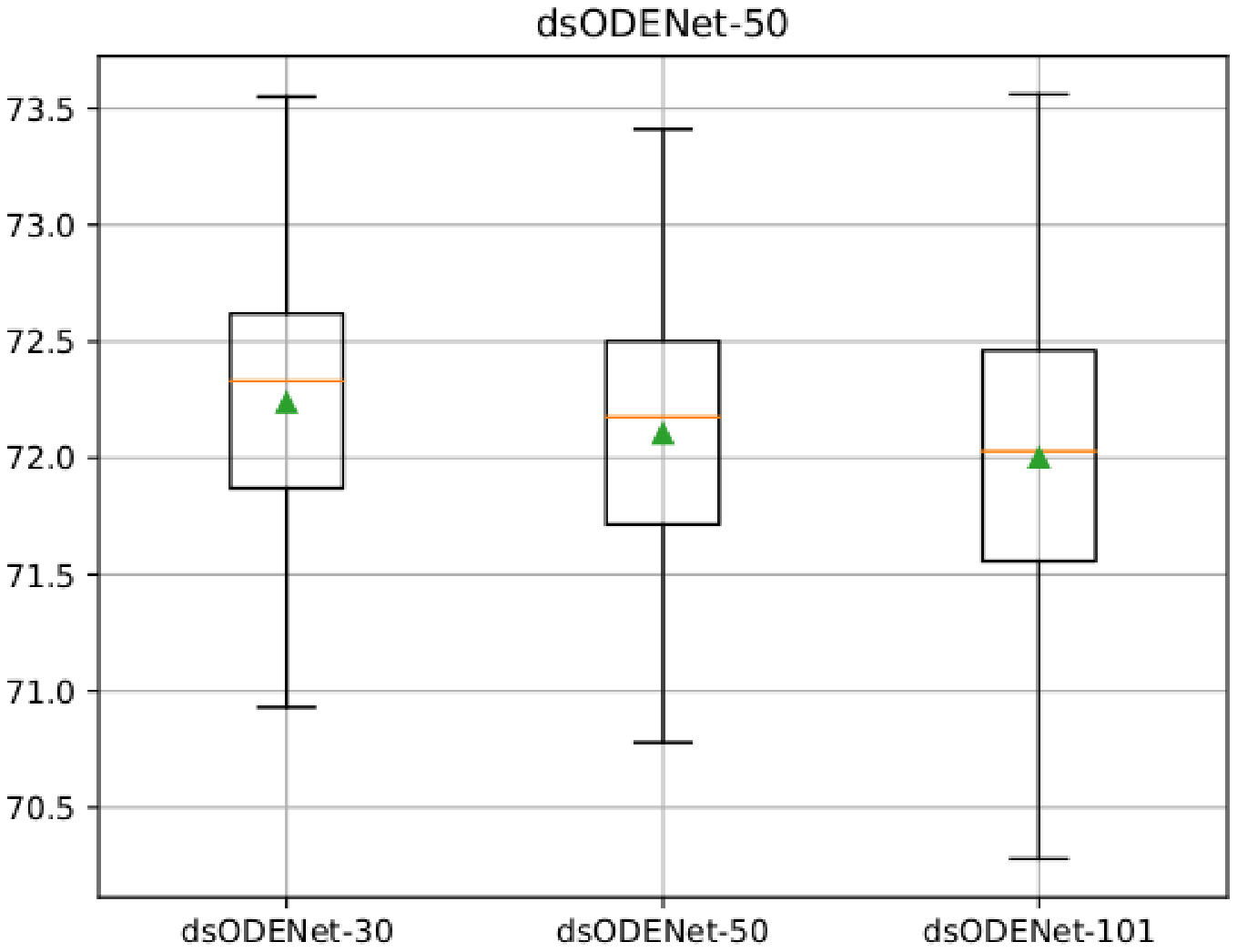}
	\caption{Accuracy of dsODENet-34, 50, and 101 using parameters
          trained as dsODENet-50 [\%]}
	\label{fig:dsodeweight50}
\end{minipage}
\end{figure}

\begin{figure}[t]
\begin{minipage}{0.49\textwidth}
	\centering
	\includegraphics[width=1\linewidth]{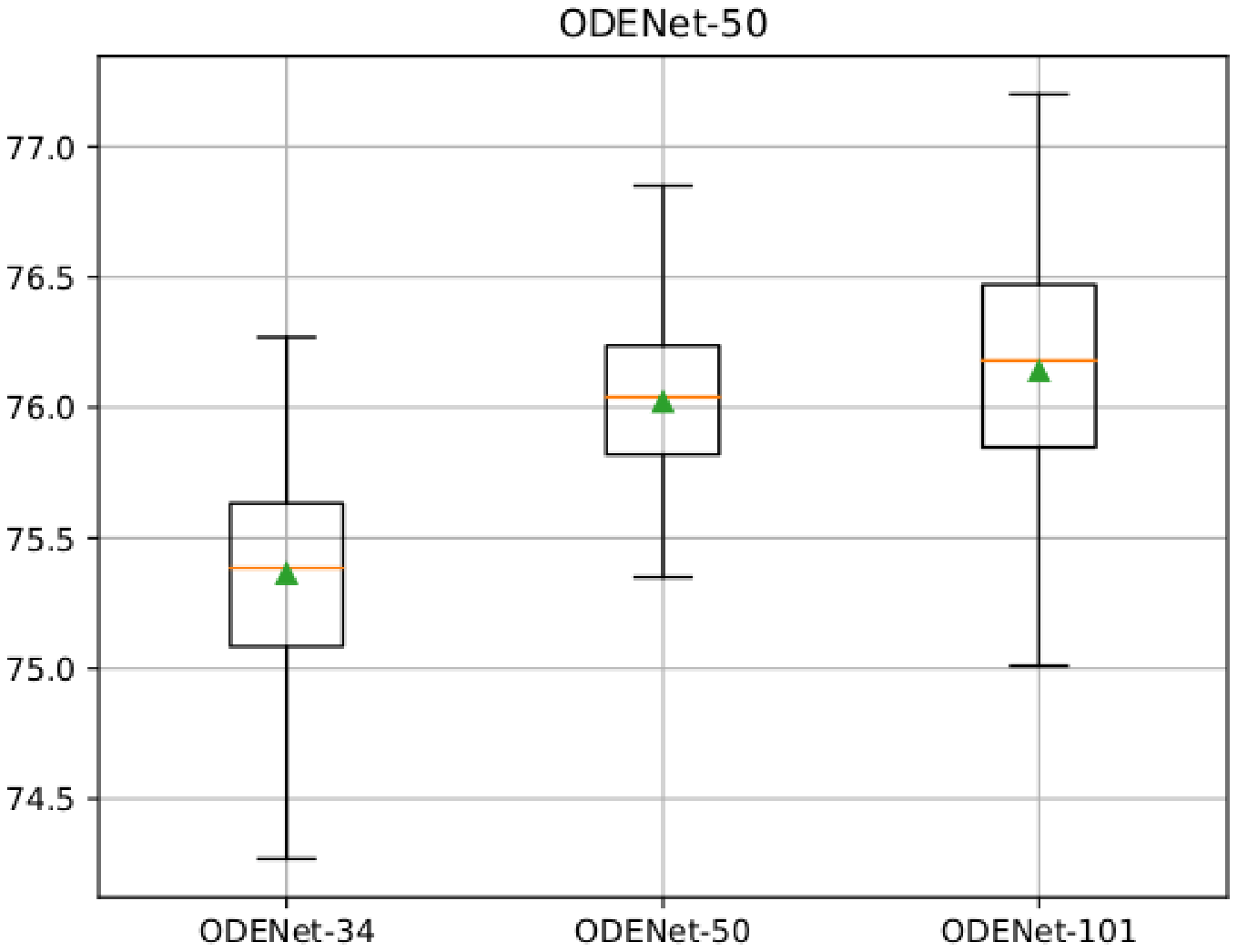}
	\caption{Accuracy of ODENet-50 using parameters trained as
          ODENet-34, 50, and 101 [\%]}
	\label{fig:odemodel50}
\end{minipage}
\begin{minipage}{0.49\textwidth}
	\centering
	\includegraphics[width=1\linewidth]{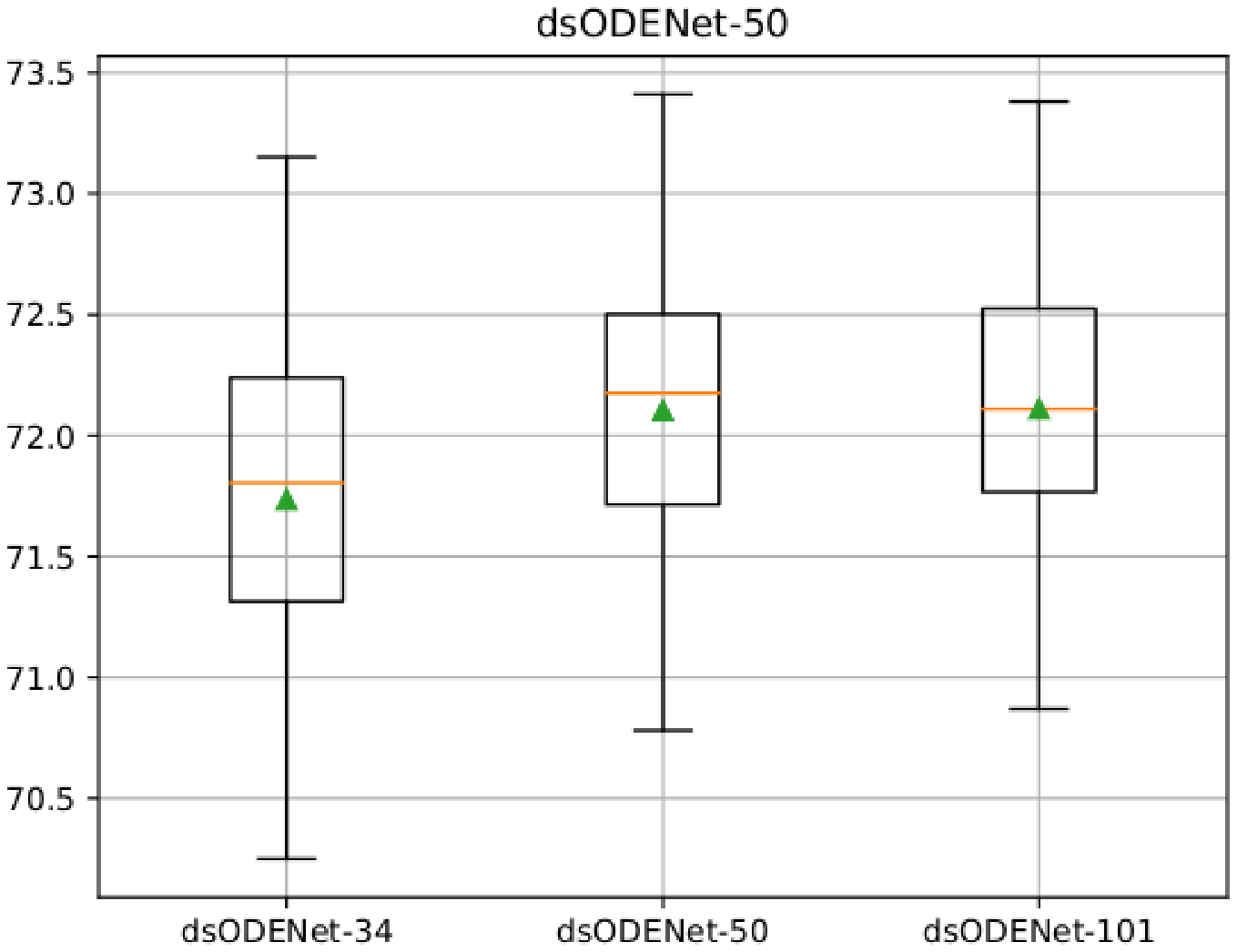}
	\caption{Accuracy of dsODENet-50 using parameters trained as
          dsODENet-34, 50, and 101 [\%]}
	\label{fig:dsmodel50}
\end{minipage}
\end{figure}

Tables \ref{table:odeDifferent} and \ref{table:dsodeDifferent} show
the inference accuracy of every combination.
Figure \ref{fig:odeweight50} shows box-plots of accuracies of
ODENet-34, 50, and 101 using weight parameters trained as ODENet-50.
Figure \ref{fig:dsodeweight50} shows those of dsODENet.
The results from Figure \ref{fig:odeweight50} and Table
\ref{table:odeDifferent} show that the accuracies are almost the same
regardless of the tested models if the trained model is the same.
The results from Figure \ref{fig:dsodeweight50} and Table
\ref{table:dsodeDifferent} also show the same tendency in dsODENet.
Figure \ref{fig:odemodel50} shows box-plots of accuracies of ODENet-50
using weight parameters trained as ODENet-34, 50, and 101, respectively.
Figure \ref{fig:dsmodel50} shows those of dsODENet-50.
The results from Figure \ref{fig:odemodel50} and Table
\ref{table:odeDifferent} show that the accuracies depend on $N$ of the
trained model.
The results from Figure \ref{fig:dsmodel50} and Table
\ref{table:dsodeDifferent} show the same tendency in dsODENet.

As mentioned above, the inference accuracies of the ODENet and
dsODENet models are reasonable even if the trained $N$ and tested $N$
are different.
This means that ODENet and dsODENet models have a weight parameter
compatibility with different depths.
Furthermore, Table \ref{table:odeDifferent} shows that the accuracies
of models that were trained as ODENet-101 are higher than those that
were trained as ODENet-34 and ODENet-50.
Table \ref{table:dsodeDifferent} also shows a similar tendency in
dsODENet.
These results indicate that using a deeper model for training can
help to enhance the accuracy in ODENet and dsODENet models.

\subsection{Federated Learning with Different Depths}\label{ssec:diffed}
In Section \ref{ssec:comp}, we showed the weight parameter compatibility
in ODENet and dsODENet models with different depths.
In this section, we examine the feasibility of federated learning
between ODENet models with different depths.
Specifically, we perform federated learning of two ODENet models from
among $N$ = 34, 50, and 101 to see if it works correctly.
In this experiment, FedAvg is used as a federated learning algorithm.
The number of clients is only 2, the number of local epochs is 5, and
the number of communication rounds is 20.
ODENet-50 is used as a global model.
The same experiment is performed for dsODENet too.

\begin{figure}[t]
\begin{minipage}{0.49\textwidth}
	\centering
	\includegraphics[width=1\linewidth]{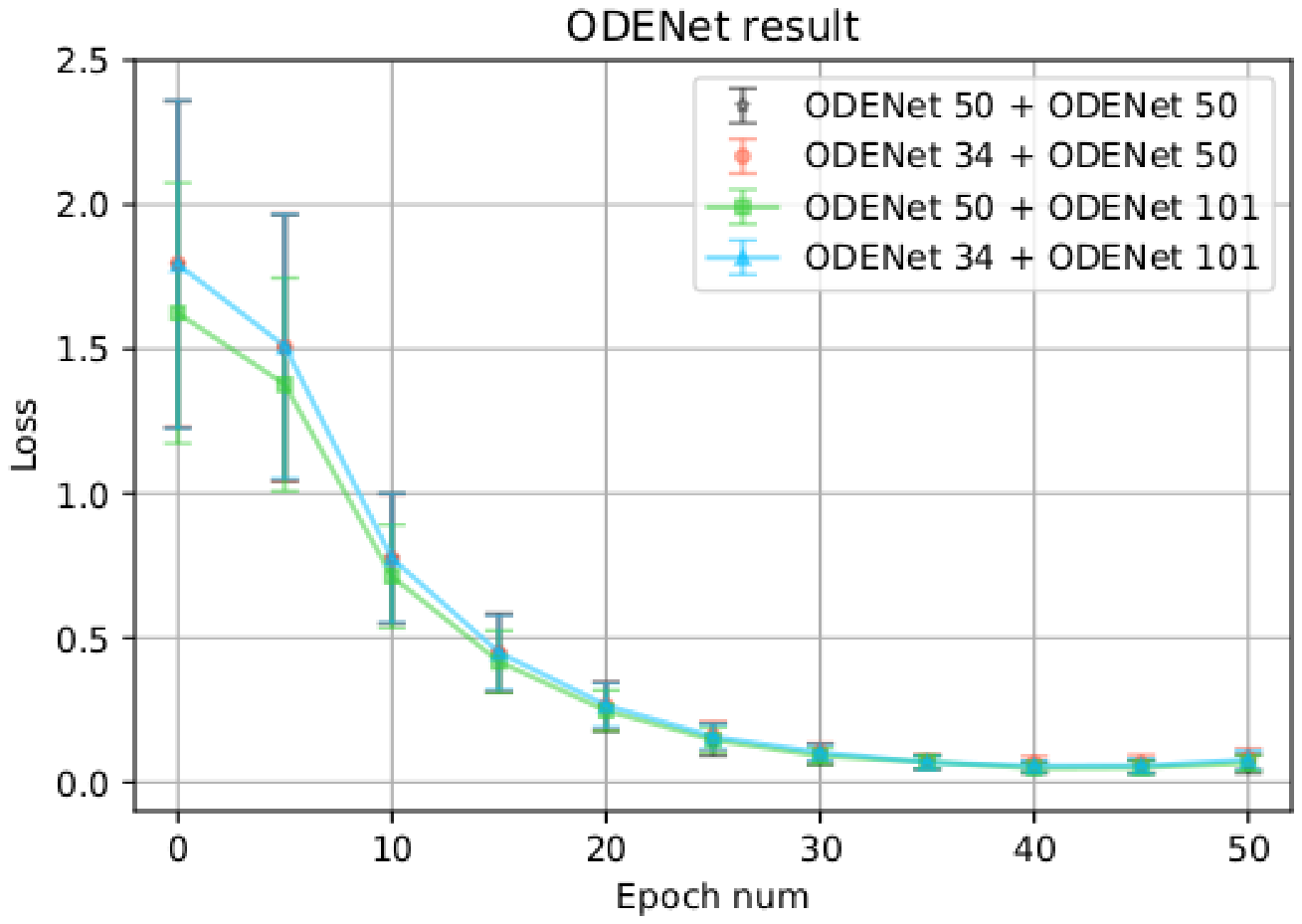}
	\caption{Training curve (epoch number vs. loss value) of ODENet}
	\label{fig:ode-loss}
\end{minipage}
\begin{minipage}{0.49\textwidth}
	\centering
	\includegraphics[width=1\linewidth]{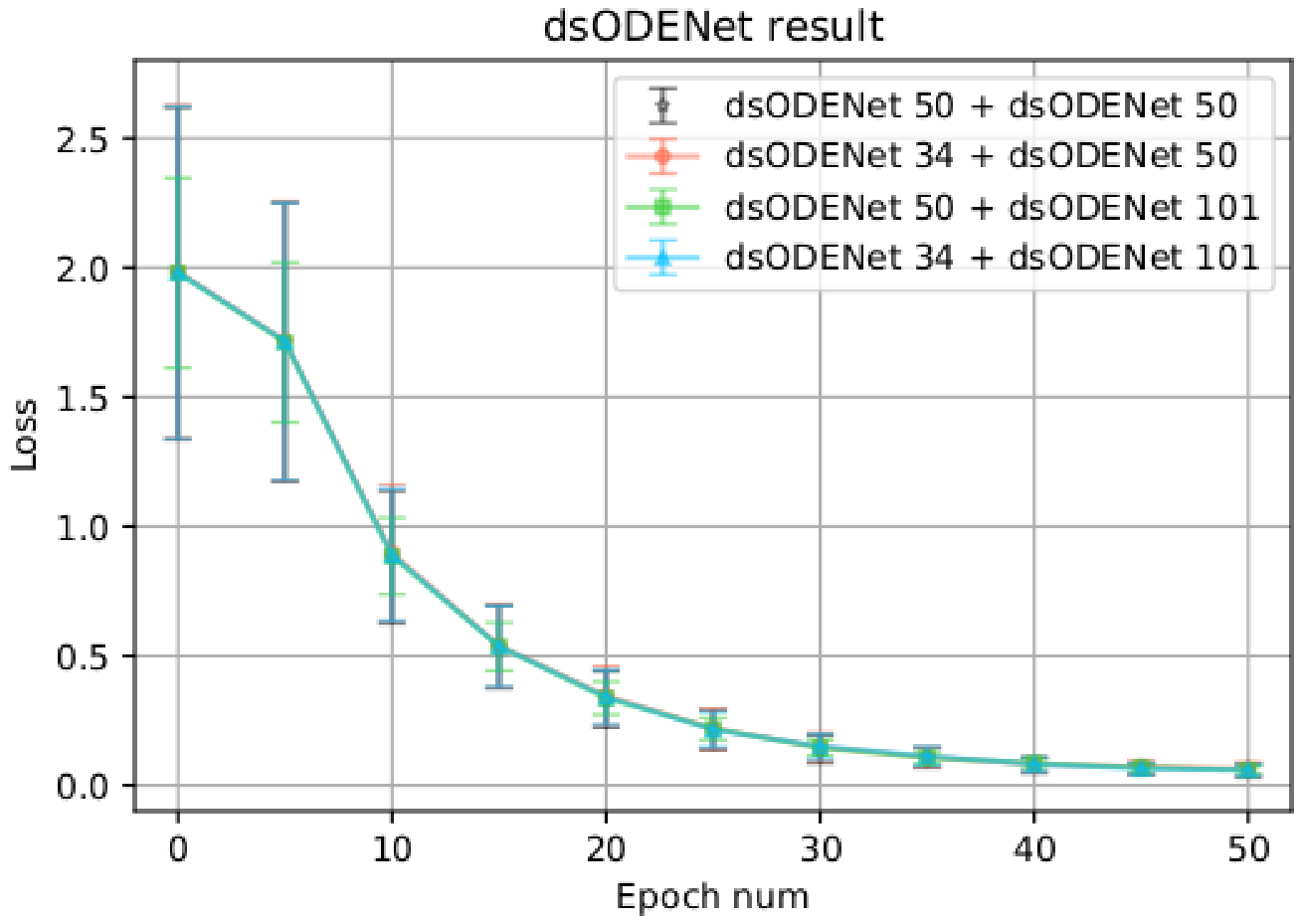}
	\caption{Training curve (epoch number vs. loss value) of dsODENet}
	\label{fig:dsode-loss}
\end{minipage}
\end{figure}

Figures \ref{fig:ode-loss} and \ref{fig:dsode-loss} show training
curves (epoch number vs. loss value) of the ODENet and dsODENet models,
respectively.
In Figure \ref{fig:ode-loss}, the black, red, green, and blue lines
show the loss values when federated learning is performed between
ODENet-50 and ODENet-50, between ODENet-34 and ODENet-50, between
ODENet-50 and ODENet-101, and between ODENet-34 and ODENet-101,
respectively.
Similarly, in Figure \ref{fig:dsode-loss}, the black, red, green, and
blue lines show the loss values when federated learning is performed
between dsODENet-50 and dsODENet-50, between dsODENet-34 and
dsODENet-50, between dsODENet-50 and dsODENet-101, and between
dsODENet-34 and dsODENet-101, respectively.
The horizontal axis of these figures represents the number of epochs,
and the vertical axis represents the loss value.
As shown, the loss values decrease as the number of epochs is
increased, and then they are converged around 50 epochs in all the
combinations.
This indicates that these model combinations of different depths can
be trained successfully.
This and previous sections demonstrated that ODENet and dsODENet are
capable	of federated learning between models with different depths.

\section{Evaluations} \label{sec:eval}
In this section, we evaluate the proposed flexible federated learning
approach that uses FedAvg as a federated learning algorithm and
ODENet, dsODENet, and ResNet as client models.
They are compared with FedDF which is a model agnostic federated
learning approach using a knowledge distillation.
Finally, the proposed approach is evaluated in terms of the model size
and communication cost.

Python 3.8.5, PyTorch 1.8.1 \cite{pytorch}, and torchvision 0.9.1 are
used for the model implementation.
A machine with Ubuntu 18.04.5 LTS (64-bit), Intel Core i7-10700K CPU @
3.8GHz, 32GB DDR4 SDRAM, and NVIDIA GeForce RTX 3090 GPU is used for
the evaluation in this paper.

\begin{figure}[t]
\begin{minipage}{0.33\textwidth}
	\centering
	\includegraphics[width=1\linewidth]{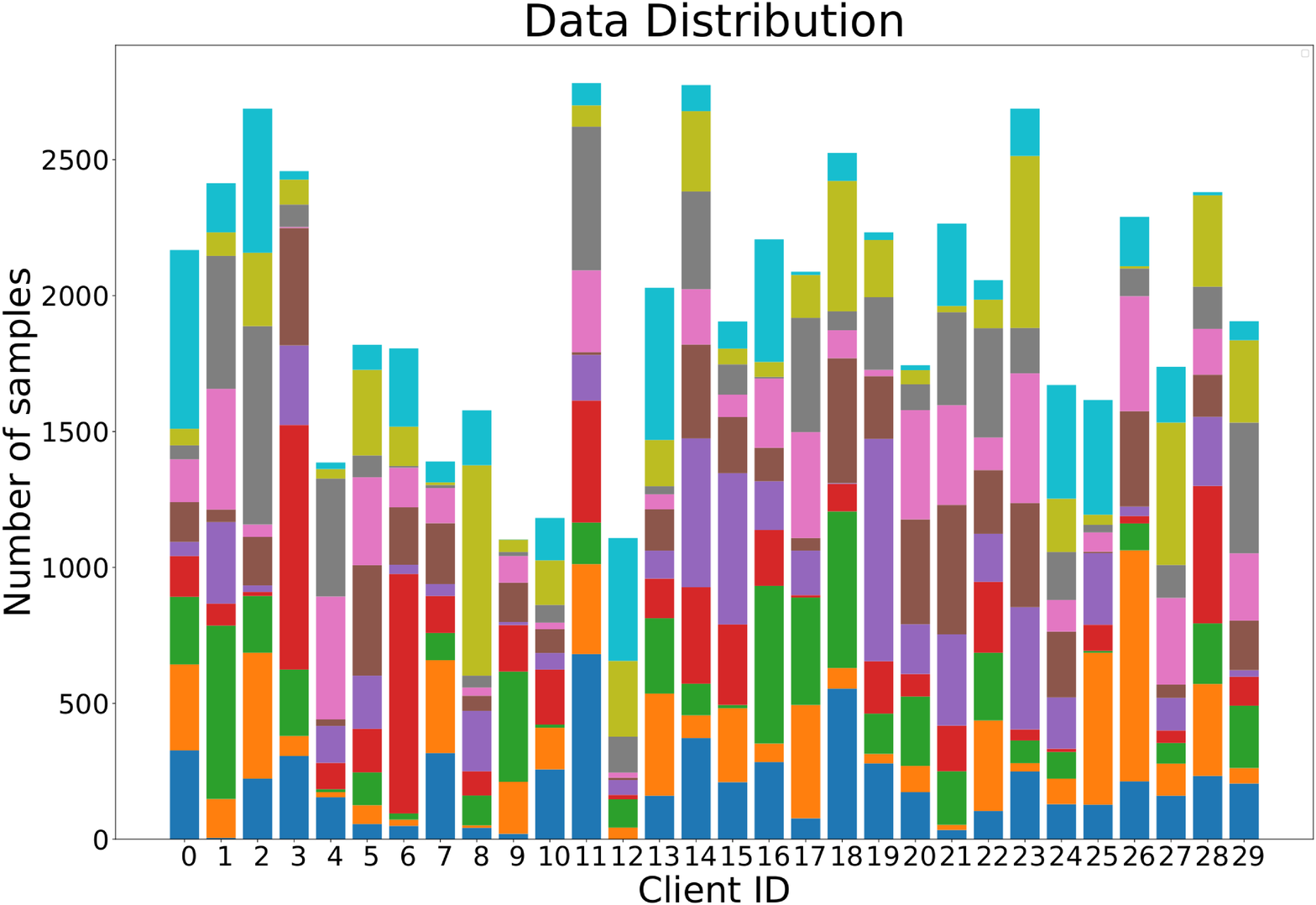}
	\caption{Data distribution using Dirichlet distribution ($\alpha$ = 1)}
	\label{fig:dist1}
\end{minipage}
\begin{minipage}{0.33\textwidth}
	\centering
	\includegraphics[width=1\linewidth]{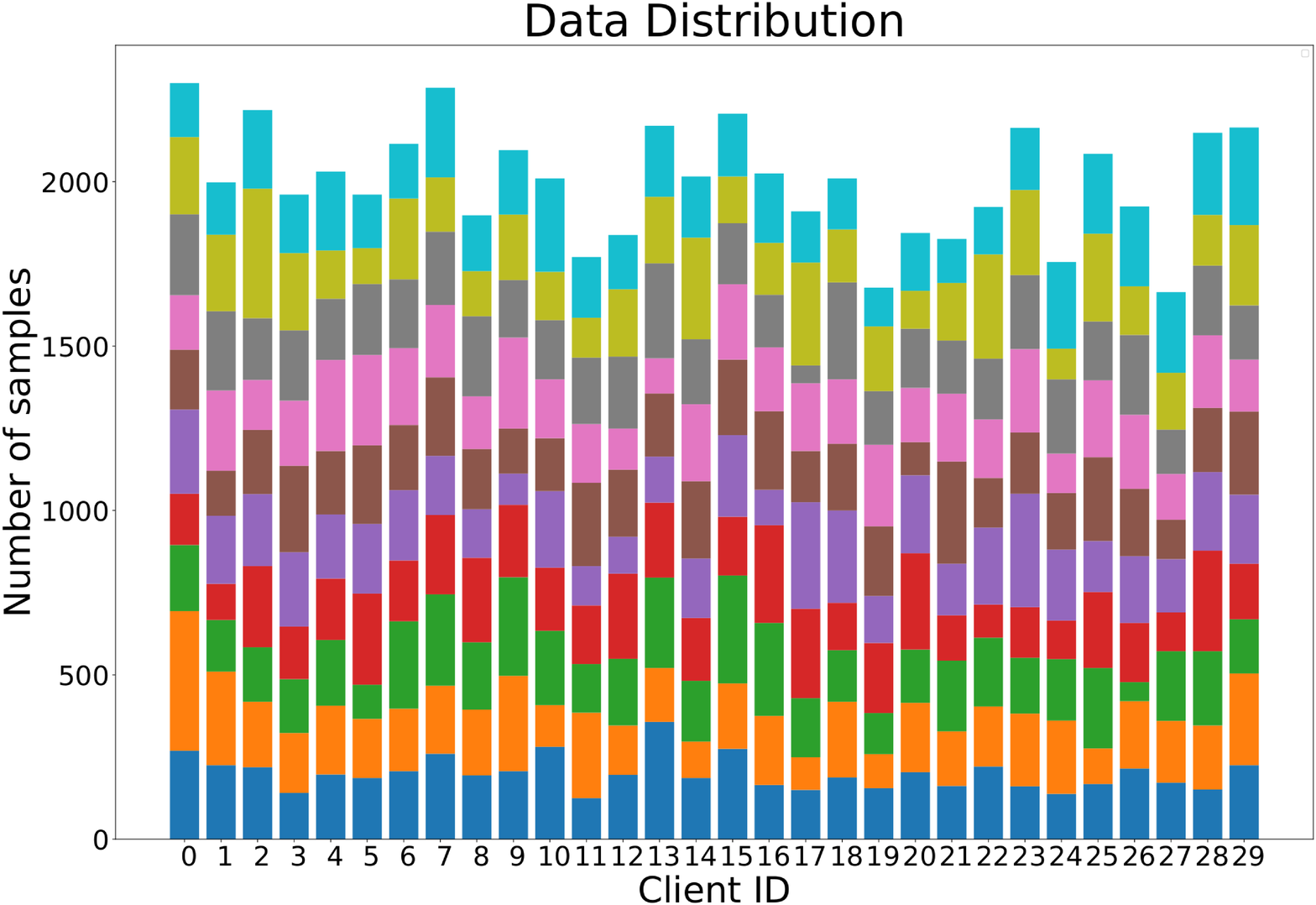}
	\caption{Data distribution using Dirichlet distribution ($\alpha$ = 10)}
	\label{fig:dist10}
\end{minipage}
\begin{minipage}{0.33\textwidth}
	\centering
	\includegraphics[width=1\linewidth]{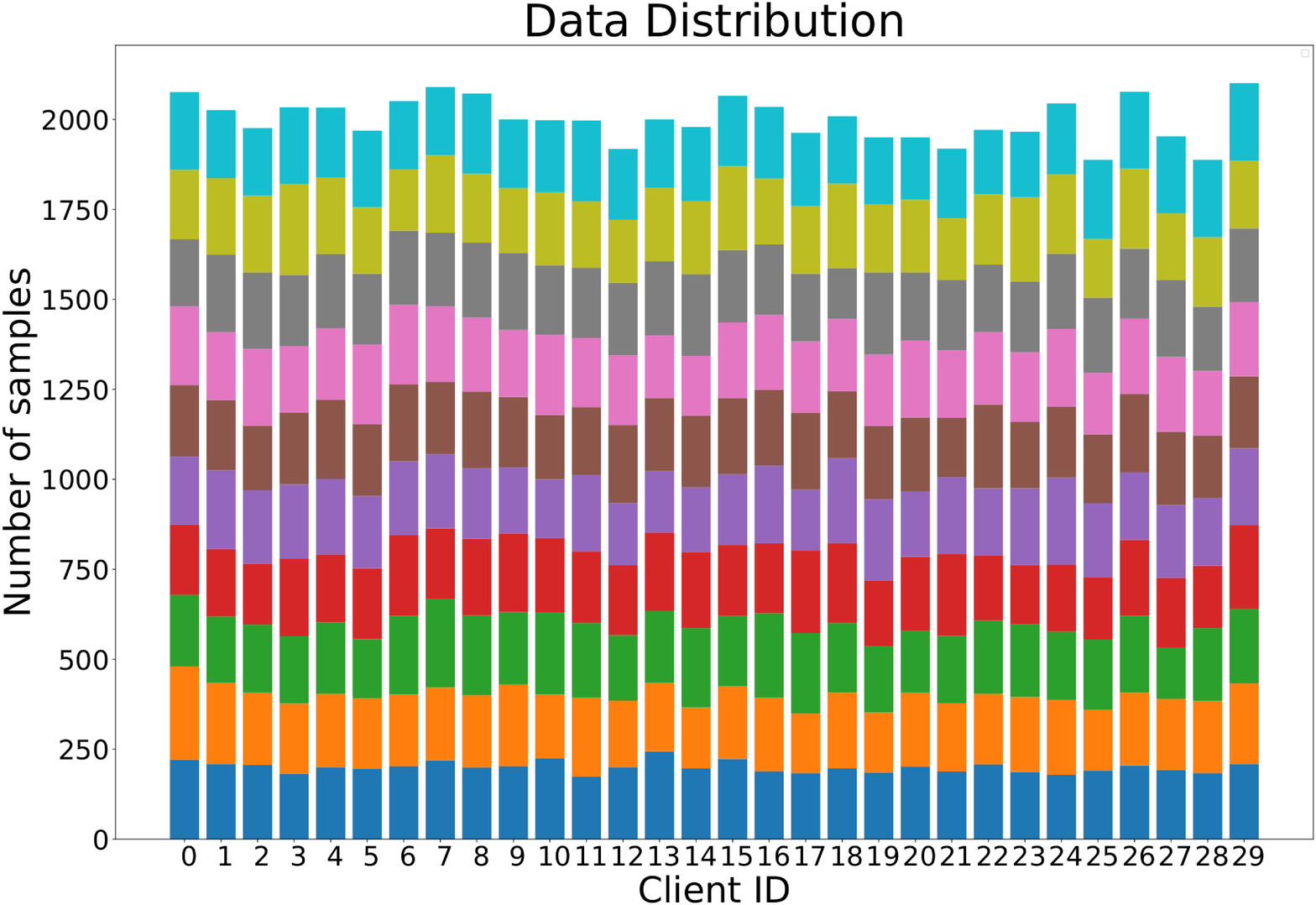}
	\caption{Data distribution using Dirichlet distribution ($\alpha$ = 100)}
	\label{fig:dist100}
\end{minipage}
\end{figure}

\subsection{Accuracy}
Although Section \ref{ssec:diffed} showed the feasibility of the
proposed federated learning, only two clients were used in the
experiments.
However, it is expected that more clients participate in a practical
federated learning scenario.
In this section, we increase the number of clients and conduct
additional experiments.

In addition, it is expected that there is a bias in the data
distribution for each client in the case of real environments.
This means that the data distribution for each client is non-iid.
In this experiment, Dirichlet distribution is thus used to make non-iid
data environments.
Dirichlet distribution is a kind of continuous multivariate
probability distributions, and its data distribution is controlled by
a vector $\alpha$.
We use $\alpha$ = 1, 10, and 100.
CIFAR-10 is used as a dataset.
Figures \ref{fig:dist1}-\ref{fig:dist100} show examples of data
distributions with $\alpha$ = 1, 10, and 100 as Dirichlet distribution
parameters, respectively.

\begin{table}[t]
	\caption{Accuracy of ODENet and dsODENet when using FedAvg and FedDF ($\alpha$ = 1)}
	\centering
	\begin{tabular}{|r|r|r|r|}
	\hline
	Model 	& Algorithm & Top5 [\%] & Top1 [\%] \\
	\hline
	ODENet	& FedAvg& 96.33 & 66.83 \\
		& FedDF & 98.23 & 73.73 \\
	\hline 
	dsODENet& FedAvg& 96.68 & 66.91 \\
		& FedDF	& 98.40 & 75.67 \\
	\hline
	\end{tabular}
	\label{table:feddf1}
\vspace{5mm}
	\caption{Accuracy of ODENet and dsODENet when using FedAvg and FedDF ($\alpha$ = 10)}
	\centering
	\begin{tabular}{|r|r|r|r|}
	\hline
	Model 	& Algorithm & Top5 [\%] & Top1 [\%] \\
	\hline
	ODENet	& FedAvg& 97.25 & 69.74 \\
		& FedDF & 98.87 & 78.55 \\
	\hline 
	dsODENet& FedAvg& 97.27 & 69.82 \\
		& FedDF	& 98.81 & 78.63 \\
	\hline
	\end{tabular}
	\label{table:feddf10}
\vspace{5mm}
	\caption{Accuracy of ODENet and dsODENet when using FedAvg and FedDF ($\alpha$ = 100)}
	\centering
	\begin{tabular}{|r|r|r|r|}
	\hline
	Model 	& Algorithm & Top5 [\%] & Top1 [\%] \\
	\hline
	ODENet	& FedAvg& 97.33 & 70.48 \\
		& FedDF & 98.78 & 78.67 \\
	\hline 
	dsODENet& FedAvg& 97.38 & 70.06 \\
		& FedDF	& 98.89 & 78.72 \\
	\hline
	\end{tabular}
	\label{table:feddf100}
\end{table}

In this experiment, the number of clients is 30, the number of local
epochs is 40, and the number of communication rounds is 100.
Among the 30 clients, the numbers of clients that use ODENet-34,
ODENet-50, and ODENet-101 are 10, 10, and 10, respectively.
ODENet-50 is used as a global model.
In Algorithm \ref{alg:federated-averaging}, $r$ is the percentage of
clients participating in the aggregation at each communication round.
In this experiment, $r$ is set to 0.2, which means that six models
are randomly selected from the 30 clients.
FedAvg is used in the proposed federated learning approach, and the
results are compared with those of FedDF.

Tables \ref{table:feddf1}-\ref{table:feddf100} show the accuracies of 
ODENet-50 and dsODENet-50 with FedAvg and FedDF when $\alpha$ = 1, 10,
and 100, respectively.
Top5 accuracies are mostly high regardless of $\alpha$ in both
the models.
Top1 accuracies are decreased when $\alpha$ is small (e.g., $\alpha$ = 1).
In this case, the data distribution is highly biased, and the
biased data distribution negatively affects the accuracy.
Although addressing the data heterogeneity is a crucial challenge in
federated learning, many studies have been conducted to overcome this
issue as mentioned in \cite{survey} and thus addressing this is beyond
the scope of this paper.

When we compare the proposed approach using FedAvg with FedDF, the
accuracies of the proposed approach are lower than those of FedDF.
This result is reasonable since FedDF introduces additional training
overheads for the knowledge distillation (e.g., prediction and
training of client models) at the server in addition to local training
at the clients while the proposed approach does not impose such
overheads as well as the conventional FedAvg based approach.
To bring the evaluation condition of FedDF closer to that of the
proposed approach, here we limit the number of samples to be used in
the knowledge distillation at the server.

Table \ref{table:datanum} shows the results of FedDF when the number 
of samples available for the knowledge distillation is varied.
$\alpha$ is set to 10 as the data distribution parameter.
The other conditions are same as those in Table \ref{table:feddf10}.
As shown in Table \ref{table:datanum}, the accuracy of FedDF increases
as the number of server-side samples used in the knowledge
distillation is increased.
When the number of the available server-side samples is small, the
accuracy of FedDF becomes close to (or lower than) the proposed approach,
as shown in Tables \ref{table:feddf10} and \ref{table:datanum}.
Please note that server-side overheads of FedDF are larger than those
of the proposed approach since FedDF requires the knowledge
distillation at the server.
In addition, FedDF requires training samples at the server; however,
uploading such samples to the server may not be easy depending on the
application due to data privacy perspective (e.g., medial image data).
On the other hand, the proposed approach can simply aggregate ODENet
based models with different iteration counts without the server-side
training nor uploading training samples to the server.


\begin{table}[t]
	\caption{Accuracy of FedDF when varying the number of samples
	used for knowledge distillation ($\alpha$ = 10)}
	\centering
	\begin{tabular}{|r|r|r|r|}
	\hline
	Model	& \# of server samples & Top5 [\%] & Top1 [\%] \\
	\hline
	ODENet	&      5 & 97.22 & 69.53 \\
		&     10 & 97.09 & 70.49 \\
		&    100 & 97.95 & 74.37 \\
		&    500 & 98.57 & 76.48 \\
		&   1000 & 98.71 & 78.32 \\
		&$\infty$& 98.87 & 78.55 \\
	\hline
	dsODENet&      5 & 97.18 & 69.52 \\
		&     10 & 97.16 & 69.82 \\
		&    100 & 97.93 & 73.88 \\
		&    500 & 98.41 & 76.89 \\
		&   1000 & 98.73 & 78.32 \\
		&$\infty$& 98.81 & 78.63 \\
		\hline
	\end{tabular}
	\label{table:datanum}
\end{table}

\subsection{Communication Size}\label{sec:commu}
Table \ref{table:trans} shows communication sizes of the ResNet,
ODENet, and dsODENet models with different depths.
Here, the communication size means the sum of the weight parameters of
convolutional layers and fully-connected layers in these models.
The communication is required between the server and clients in each
communication round.
The size is measured by using torchsummary of PyTorch which is a tool
that reports the parameter size information.

\begin{table}[t]
	\caption{Communication size of ResNet, ODENet, and dsODENet}
	\centering
	\begin{tabular}{|r|r|r|}
	  \hline
	  Model & \# of parameters transferred & Size [MB] \\
	  \hline 
	  ResNet-34 & 21,780,648 & 83.09\\
	  ResNet-50 & 25,505,232 & 97.29\\
	  ResNet-101 & 44,447,912 & 169.55\\
	  \hline
	  ODENet-34 & 1,937,034 & 7.39\\
	  ODENet-50 & 1,937,034 & 7.39\\
	  ODENet-101 & 1,937,034& 7.39\\
	  \hline
	  dsODENet-34 & 1,249,381 & 4.76\\
	  dsODENet-50 & 1,249,381 & 4.76\\
	  dsODENet-101 & 1,249,381 & 4.76\\
	  \hline
	\end{tabular}
	\label{table:trans}
\end{table}

Table \ref{table:trans} shows that both the ODENet and dsODENet models
reduce the communication sizes compared with the corresponding ResNet
model.
Compared with ResNet, the communication size of ODENet-50 is
7.6\% of the original ResNet model.
In the case of ResNet and dsODENet, the communication size of
dsODENet-50 is 4.9\% of the ResNet model.
These results show that the use of ODENet and dsODENet can
significantly reduce the communication size between the server and
clients.

The communication size increases as $N$ is increased in ResNet.
On the other hand, communication sizes are constant regardless of $N$
in the cases of ODENet and dsODENet.
This is because the number of physically-stacked blocks is the same in
ODENet and dsODENet even if $N$ is different, and only the number of
iterations of each block is different.
Please note that, since the model size is small in ODENet and
dsODENet, the required memory capacity can also be reduced by these
models compared with the original ResNet model.


\section{Conclusions}\label{sec:conc}
In this paper, we proposed a flexible federated learning approach that
can aggregate models with different iteration counts by utilizing
ODENet and dsODENet as federated learning models.
We demonstrated that these models with different iteration counts can
be aggregated correctly (i.e., having the weight compatibility) in the
cases of ODENet and dsODENet.
Then, the proposed approach simply using FedAvg was compared with
FedDF in terms of the accuracy.
The experiment results showed that the higher accuracy of FedDF come
from additional knowledge distillation overheads at the server.
On the other hand, the proposed approach can simply aggregate models
with different iteration counts without the server-side training nor
uploading training samples to the server.
In addition, ODENet and dsODENet were evaluated in terms of the
model and communication sizes.
Compared with ResNet-50, ODENet-50 and dsODENet-50 successfully
reduced the communication sizes by 92.4\% and by 95.1\%, respectively.
These results showed that our approach can significantly reduce the
communication overhead while enabling the aggregation of models with
different iteration counts.

As a future work, we will evaluate the feasibility of federated
learning with ANODE \cite{anode} which is a Neural ODE based approach
that utilizes a checkpointing method.
We are also planning to improve accuracy when $\alpha$ is small.
We will combine our approach with state-of-the-art federated
learning algorithms.


\begin{thebibliography}{10}

\bibitem{Federated}
Brendan McMahan, Eider Moore, Daniel Ramage, Seth Hampson, and Blaise~Aguera
  y~Arcas.
\newblock {Communication-Efficient Learning of Deep Networks from Decentralized
  Data}.
\newblock In {\em Proceedings of the International Conference on Artificial
  Intelligence and Statistics (AISTATS)}, pages 1273--1282, April 2017.

\bibitem{odenet}
Ricky T.~Q. Chen, Yulia Rubanova, Jesse Bettencourt, and David Duvenaud.
\newblock {Neural Ordinary Differential Equations}.
\newblock In {\em Proceedings of the Annual Conference on Neural Information
  Processing Systems (NeurIPS)}, pages 6572--6583, December 2018.

\bibitem{resnet}
Kaiming He, Xiangyu Zhang, Shaoqing Ren, and Jian Sun.
\newblock {Deep Residual Learning for Image Recognition}.
\newblock In {\em Proceedings of the IEEE Conference on Computer Vision and
  Pattern Recognition (CVPR)}, pages 770--778, June 2016.

\bibitem{dsode}
Hiroki Kawakami, Hirohisa Watanabe, Keisuke Sugiura, and Hiroki Matsutani.
\newblock {A Low-Cost Neural ODE with Depthwise Separable Convolution for Edge
  Domain Adaptation on FPGAs}.
\newblock {\em IEICE Transactions on Information and Systems},
  E106-D(7):1186--1197, July 2023.

\bibitem{mobilenets}
Andrew~G. Howard, Menglong Zhu, Bo~Chen, Dmitry Kalenichenko, Weijun Wang,
  Tobias Weyand, Marco Andreetto, and Hartwig Adam.
\newblock {MobileNets: Efficient Convolutional Neural Networks for Mobile
  Vision Applications}.
\newblock {arXiv Preprint arXiv:1704.04861}, April 2017.

\bibitem{hoshinocw}
Yuto Hoshino, Hiroki Kawakami, and Hiroki Matsutani.
\newblock {Communication Size Reduction of Federated Learning based on Neural
  ODE Model}.
\newblock In {\em Proceedings of the International Symposium on Computing and
  Networking (CANDAR) Workshops}, pages 55--61, November 2022.

\bibitem{survey}
Qinbin Li, Zeyi Wen, Zhaomin Wu, Sixu Hu, Naibo Wang, Yuan Li, Xu~Liu, and
  Bingsheng He.
\newblock {A Survey on Federated Learning Systems: Vision, Hype and Reality for
  Data Privacy and Protection}.
\newblock {\em IEEE Transactions on Knowledge and Data Engineering},
  35(4):3347--3366, April 2021.

\bibitem{fedprox}
Tian Li, Anit~Kumar Sahu, Manzil Zaheer, Maziar Sanjabi, Ameet Talwalkar, and
  Virginia Smith.
\newblock {Federated Optimization in Heterogeneous Networks}.
\newblock {arXiv Preprint arXiv:1812.06127}, April 2020.

\bibitem{scaf}
Sai~Praneeth Karimireddy, Satyen Kale, Mehryar Mohri, Sashank Reddi, Sebastian
  Stich, and Ananda~Theertha Suresh.
\newblock {SCAFFOLD: Stochastic Controlled Averaging for Federated Learning}.
\newblock In {\em Proceedings of the International Conference on Machine
  Learning (ICML)}, pages 5132--5143, July 2020.

\bibitem{perfed}
Alireza Fallah, Aryan Mokhtari, and Asuman Ozdaglar.
\newblock {Personalized Federated Learning with Theoretical Guarantees: A
  Model-Agnostic Meta-Learning Approach}.
\newblock In {\em Proceedings of the Annual Conference on Neural Information
  Processing Systems (NeurIPS)}, pages 3557--3568, December 2020.

\bibitem{apfl}
Yuyang Deng, Mohammad~Mahdi Kamani, and Mehrdad Mahdavi.
\newblock {Adaptive Personalized Federated Learning}.
\newblock {arXiv Preprint arXiv:2003.13461}, March 2020.

\bibitem{fedhenn}
Disha Makhija, Xing Han, Nhat Ho, and Joydeep Ghosh.
\newblock {Architecture Agnostic Federated Learning for Neural Networks}.
\newblock In {\em Proceedings of the International Conference on Machine
  Learning (ICML)}, pages 14860--14870, July 2022.

\bibitem{feddf}
Tao Lin, Lingjing Kong, Sebastian~U. Stich, and Martin Jaggi.
\newblock {Ensemble Distillation for Robust Model Fusion in Federated
  Learning}.
\newblock {arXiv Preprint arXiv:2006.07242}, March 2021.

\bibitem{cifar10}
Alex Krizhevsky.
\newblock {Learning Multiple Layers of Features from Tiny Images}, April 2009.

\bibitem{pytorch}
Adam Paszke, Sam Gross, Francisco Massa, Adam Lerer, James Bradbury, Gregory
  Chanan, Trevor Killeen, Zeming Lin, Natalia Gimelshein, Luca Antiga, Alban
  Desmaison, Andreas Kopf, Edward Yang, Zachary DeVito, Martin Raison, Alykhan
  Tejani, Sasank Chilamkurthy, Benoit Steiner, Lu~Fang, Junjie Bai, and Soumith
  Chintala.
\newblock {PyTorch: An Imperative Style, High-Performance Deep Learning
  Library}.
\newblock In {\em Proceedings of the Annual Conference on Neural Information
  Processing Systems (NeurIPS)}, pages 8024--8035, December 2019.

\bibitem{anode}
Amir Gholaminejad, Kurt Keutzer, and George Biros.
\newblock {ANODE: Unconditionally Accurate Memory-Efficient Gradients for
  Neural ODEs}.
\newblock In {\em Proceedings of the International Joint Conference on
  Artificial Intelligence (IJCAI)}, pages 730--736, August 2019.

\end{thebibliography}

\end{document}